\documentclass[runningheads]{llncs}
\usepackage[table]{xcolor}
\usepackage{times}
\usepackage{xspace}
\usepackage{pifont}
\usepackage{enumitem}

\usepackage{graphicx}
\usepackage{todonotes}
\usepackage{amsmath}
\usepackage{amsfonts}
\usepackage{amssymb}
\usepackage{tikz}
\usepackage{changepage}
\usepackage{booktabs}
\usepackage{multirow}
\usepackage{tabularx}

\usepackage{lipsum}
\newcommand\blfootnote[1]{%
  \begingroup
  \renewcommand\thefootnote{}\footnote{#1}%
  \addtocounter{footnote}{-1}%
  \endgroup
}
\begin{document}
\title{Graph Attention Networks with Positional Embeddings}
\newcommand{\model}{GAT-POS\xspace}
\author{Liheng Ma \inst{1,2} \and 
Reihaneh Rabbany \inst{1,2} \and 
Adriana Romero-Soriano \inst{1,3}
\scriptsize{\email{liheng.ma@mail.mcgill.ca\and rrabba@cs.mcgill.ca\and adrianars@fb.com }}
}
\institute{SCS McGill University, Montreal, Canada \and Mila, Montreal, Canada\and  Facebook AI Research, Montreal, Canada
}

\authorrunning{L. Ma et al.}
%
\maketitle              
\begin{abstract}
Graph Neural Networks (GNNs) are deep learning methods which provide the current state of the art performance in node classification tasks.
GNNs often assume homophily -- neighboring nodes having similar features and labels--, and therefore may not be at their full potential when dealing with non-homophilic graphs.
In this work, we focus on addressing this limitation and enable Graph Attention Networks (GAT), a commonly used variant of GNNs, to explore the structural information within each graph locality. 
Inspired by the positional encoding in the Transformers, we propose a framework, termed Graph Attentional Networks with Positional Embeddings (GAT-POS), to enhance GATs with positional embeddings which capture structural and positional information of the nodes in the graph.
In this framework, the positional embeddings are learned by a model predictive of the graph context, plugged into an enhanced GAT architecture, which is able to leverage both the positional and content information of each node.
The model is trained jointly to optimize for the task of node classification as well as the task of predicting graph context.
Experimental results show that GAT-POS reaches remarkable improvement compared to strong GNN baselines and recent structural embedding enhanced GNNs on non-homophilic graphs.

\end{abstract}
\section{Introduction} 
\label{sec:intro}

The use of graph-structured data is ubiquitous in a wide range of applications, from social networks, to biological networks, telecommunication networks, 3D vision or physics simulations. The recent years have experienced a surge in graph representation learning methods, with Graph Neural Networks (GNNs) currently being at the forefront of many application domains. Recent advances in this direction are often categorized as spectral approaches and spatial approaches. \blfootnote{Published as a conference paper at PAKDD 2021}

Spectral approaches define the convolution operator in the spectral domain and therefore capture the structural information of the graph~\cite{bruna2014spectral}. However, such approaches require computationally intense operations and yield filters which may not be localized in the spatial domain.
A number of works have been proposed to localize spectral filters and approximate them for computation efficiency~\cite{defferrard2016convolutional,kipf2016semi,lanczos1950iteration}.
However, spectral filters based on the eigenbasis of the graph Laplacian depend on the graph structure, and thus cannot be directly applied to new graph structures, which limits their performance in the inductive setting.

Spatial approaches directly define a spatially localized operator on the localities of the graph (i.e., neighborhoods), and as such are better suited to generalize to new graphs. 
Most spatial approaches aggregate node features within the localities, and then mix the aggregated features channel-wise through a linear transformation \cite{chollet2017xception,hamilton2017inductive,xu2018how,li2015gated,scarselli2008graph}.

The spatial aggregation operator has been implemented as mean-pooling or max-pooling in
~\cite{hamilton2017inductive}, and sum-pooling in \cite{xu2018how,li2015gated,scarselli2008graph}.
In order to increase the expressive power of spatial approaches, previous works have defined the aggregation operators with adaptive kernels in which the coefficients are parameterized as a function of node features. 
For instance, MoNets~\cite{monti2017geometric} define the adaptive filters based on a Gaussian mixture model formulation, and graph attention networks~(GATs)~\cite{velivckovic2017graph} introduce a content-based attention mechanisms to parameterize the coefficients of the filter.

Among them, GATs \cite{velivckovic2017graph} have become widely used and shown great performance in node classification~\cite{liao2019lanczosnet,vashishth2019confidence,gao2019graphunet}. However, the GATs' adaptive filter computation  
is based on node content exclusively and attention mechanisms cannot inherently capture the structural dependencies among entities at their input~\cite{vaswani2017attention,lee2019set}, considering them as a set structure. Therefore, GAT's filters cannot fully explore the structural information of the graph either. 
This may put the GAT framework at a disadvantage when learning on non-homophilic graph datasets, in which edges merely indicate the interaction between two nodes instead of their similarity~\cite{pei2020geom}.
Compared to homophilic graphs in which edges indicate similarity between the connected nodes, non-homophilic graphs are more challenging and higher-level structural patterns might be required to learn the node labels.

In sequence-based~\cite{vaswani2017attention,shaw2018self,dai2019transformerxl}, tree-based~\cite{wang2019structuralpos} and image-based \cite{cordonnier2019relationship,zhao2020exploring} tasks, the lack of structural information leveraged by the attention mechanisms has been remedied by introducing handcrafted or learned positional encodings~\cite{vaswani2017attention,shaw2018self,wang2019structuralpos,dai2019transformerxl,cordonnier2019relationship,zhao2020exploring}, resulting in improved performances.

Inspired by these positional encodings, and to improve learning on non-homophilic graphs, we aim to enhance GATs with structural information. Therefore, we propose a framework, called \emph{Graph Attention Networks with Positional Embeddings} (GAT-POS), which leverages both positional information and node content information in the attention computation. 
More precisely, we modify the graph attention layers to incorporate a positional embedding for each node, produced by an positional embedding model predictive of the graph context. 
Our GAT-POS model is trained end-to-end to optimize for a node classification task while simultaneously learning the node positional embeddings. 
Our results on non-homophilic datasets highlight the potential of the proposed approach, notably outperforming GNN baselines as well as the recently introduced Geom-GCN~\cite{pei2020geom}, a method tailored to perform well on non-homophilic datasets. Moreover, as a sanity check, we validate the proposed framework on standard homophilic datasets and demonstrate that GAT-POS can reach a comparable performance to GNN baselines. To summarize, the contributions of this paper are: 

\begin{itemize}
\vspace{-0.25cm}
    \item We propose GAT-POS, a novel framework which enhances GAT with positional embeddings for learning on non-homophilic graphs. The framework enhances the graph attentional layers to leverage both node content and positional information.
    \item We develop a joint training scheme for the proposed GAT-POS to support end-to-end training and learn positional embeddings tuned for the supervised task. 
    \item We show experimentally that GAT-POS significantly outperforms other baselines on non-homophilic datasets.
   
\end{itemize}

\section{Method}

In this paper, we consider the semi-supervised node classification task, and follow the problem setting of GCNs~\cite{kipf2016semi} and GATs~\cite{velivckovic2017graph}. 
Let $G=(\mathcal{V}, \mathcal{E})$ be an undirected and unweighted graph, with a set of $N$ nodes $\mathcal{V}$ and a set of edges $\mathcal{E}$. Each node $v\in \mathcal{V}$ is represented with a $D$-dimensional feature vector
 $\mathbf{x}_v \in \mathbb{R}^D$. Similarly, each node has an associated label represented with a one-hot vector $\mathbf{y}_v \in \{0, 1\}^C$, where $C$ is the number of classes.

The edges in $\mathcal{E}$ are represented as an adjacency matrix $\mathbf{A}=\{0,1\}^{N\times N}$, in which $\mathbf{A}_{vu} = 1 \text{\ iff\ } (v,u) \in \mathcal{E}$.
The neighborhood of a node is defined as $\mathcal{N}(v)=\{u|(v,u) \in \mathcal{E}\}$.
 
In the rest of the section, we will present our proposed GAT-POS framework, and detail the implementation of each one of its components.  

\subsection{The GAT-POS Model}

Multiple previous works have been proposed to incorporate positional embeddings in the attention mechanisms for sequences and grids~\cite{vaswani2017attention,dai2019transformerxl,shaw2018self,wang2019structuralpos,zhao2020exploring,cordonnier2019relationship}.
However, the structure in a graph is more complicated and the positional encoding for these earlier works cannot be directly generalized to the graphs. Therefore, we propose our framework, Graph Attention Networks with Positional Embeddings (GAT-POS) to incorporate the positional embeddings in the attention mechanisms of GATs. In particular, we propose to learn the positional embeddings via an embedding model with an unsupervised objective $\mathcal{L}_u$, termed \emph{positional embedding model}, which allows the positional embeddings to capture richer positional and structural information of the graph. We provide a detailed comparison with the previous works in Section \ref{sec:related_work}. 

In order to support end-to-end training, we propose a main-auxiliary architecture where the positional embedding model is plugged into GATs, inspired by the works of ~\cite{weston2008deep,yang2016revisiting}.
With this architecture, the supervised task of the GATs enhanced with positional embeddings, and the unsupervised task of the positional embedding model are trained jointly.
Consequently, besides supporting end-to-end training, the positional embedding model can learn embeddings not only predictive of the graph context, but also beneficial to the supervised task.

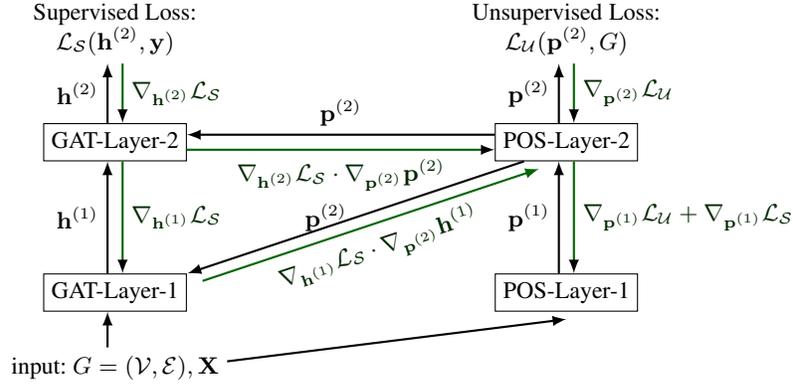
\begin{figure}[th!]
\vspace{-0.5cm}
    \centering
\begin{tikzpicture}
    \tikzstyle{layer} = [rectangle, draw=black!100, fill=white!20, align=center];
    \tikzstyle{ret} = [rectangle, draw=white!200, fill=white!20, align=center];
    \tikzstyle{eq} = [rectangle, draw=white!200, fill=white!20, align=center, opacity=0, text opacity=1];
    \tikzstyle{fwv} = [-latex, draw, thick, transform canvas={xshift=-1mm}];  
    \tikzstyle{fwh} = [-latex, draw, thick, transform canvas={yshift=1mm}];  
    \tikzstyle{bwv} = [latex-, draw=black!60!green, thick, transform canvas={xshift=1mm}];
    \tikzstyle{bwh} = [latex-, draw=black!60!green, thick, transform canvas={yshift=-1mm}];
    \node (input) at (0, 0) [ret]{input: $G=(\mathcal{V}, \mathcal{E}), \mathbf{X}$};
    \node (gat1) at (0, 1) [layer]{GAT-Layer-1};
    \node (emb1) at (6, 1) [layer]{POS-Layer-1};
    \node (gat2) at (0, 3) [layer]{GAT-Layer-2};
    \node (emb2) at (6, 3) [layer]{POS-Layer-2};
    \node (suploss) at (0, 4.5) [ret]{Supervised Loss:\\
    $\mathcal{L}_{\mathcal{\mathcal{S}}}(\mathbf{h}^{(2)}, \mathbf{y})$};
    \node (unsuploss) at (6, 4.5) [ret]{Unsupervised Loss:\\ $\mathcal{L}_{\mathcal{U}}(\mathbf{p}^{(2)}, G)$};

    \path [fwv] (input) --  (gat1);
    \path [-latex, draw, thick, transform canvas={yshift=-1mm}] (input) -- (emb1.south); 
    \path [fwv] (gat1) -- node[left, ]{$\mathbf{h}^{(1)}$} (gat2);
    \path [fwv] (emb1) -- node[left]{$\mathbf{p}^{(1)}$} (emb2);
    \path [-latex, draw, thick, transform canvas={xshift=2mm}] (emb2) -- node[above, left]{$\mathbf{p}^{(2)}$} (gat1);
    \path [fwh] (emb2) -- node[above]{$\mathbf{p}^{(2)}$} (gat2);
    \path [fwv] (gat2) -- node[left]{$\mathbf{h}^{(2)}$}  (suploss);
    \path[fwv] (emb2) -- node[left]{$\mathbf{p}^{(2)}$} (unsuploss);
    \path [bwv] (gat1) -- node[right, black!85!green]{$\nabla_{\mathbf{h}^{(1)}} \mathcal{L}_{\mathcal{S}}$} (gat2);
    \path [bwv] (emb1) -- node[right, black!85!green]{$\nabla_{\mathbf{p}^{(1)}} \mathcal{L}_{\mathcal{U}}+ \nabla_{\mathbf{p}^{(1)}}\mathcal{L}_{\mathcal{S}}$} (emb2);
    \path [latex-, draw, thick, transform canvas={xshift=4mm, yshift=-1mm}, black!60!green] (emb2) -- node[below, sloped, black!85!green]{$\nabla_{\mathbf{h}^{(1)}} \mathcal{L}_{\mathcal{S}}\cdot\nabla_{\mathbf{p}^{(2)}} \mathbf{h}^{(1)}$} (gat1);
    \path [bwh] (emb2) -- node[below, black!85!green]{$\nabla_{\mathbf{h}^{(2)}} \mathcal{L}_{\mathcal{S}}\cdot \nabla_{\mathbf{p}^{(2)}} \mathbf{p}^{(2)}$} (gat2);
    \path [bwv] (gat2) -- node[right, black!85!green]{$\nabla_{\mathbf{h}^{(2)}} \mathcal{L}_{\mathcal{S}}$}  (suploss);
    \path [bwv] (emb2) -- node[right, black!85!green]{$\nabla_{\mathbf{p}^{(2)}} \mathcal{L}_{\mathcal{U}}$} (unsuploss);
    \end{tikzpicture}
    \caption{A Demonstration of GAT-POS Architecture (the subscriptions are dropped for simplicity)}
    \label{fig:gat_pos}
    \vspace{-0.5cm}
\end{figure}

Fig.~\ref{fig:gat_pos} provides an overview of the GAT-POS architecture, where each rectangle denotes the operation of a layer; the black arrows denote forward propagation and the green arrows denote backpropagation.

For the node classification tasks, the supervised loss function is the cross-entropy error over the set of labeled examples observed during training $\mathcal{Y}_L$, as follows,
\begin{equation}
    \mathcal{L}_\mathcal{S} (\{\hat{\mathbf{y}}_v\}_{v \in \mathcal{Y}_L},\{\mathbf{y}_v\}_{v \in \mathcal{Y}_L}) = - \sum_{v \in \mathcal{Y}_L} \mathbf{y}_v^\intercal \log \hat{\mathbf{y}}_v
\end{equation}
where $\hat{\mathbf{y}}_v$ is node $v$'s predicted label. The unsupervised task is designed to guide the positional embedding model to capture information beneficial to the supervised task.

In the following subsection, we will introduce a particular implementation of GAT-POS.
However, it is worth noting that our framework is agnostic to the particular setup of the enhanced graph attentional layer, the architecture of the positional embedding model, the choice of the unsupervised objective as well as the way the main and auxiliary architectures are connected.

\subsection{Positional Embedding Enhanced Graph Attentional Layer}

We extend the graph attentional layer of \cite{velivckovic2017graph} to leverage the node embeddings extracted from the GAT-POS positional embedding model when computing the attention coefficients. 
We modify the graph attentional layer to consider positional embeddings in the attention scores computation. 

In particular, our positional embedding enhanced graph attentional layer transforms a vector of node features $\mathbf{h}_v \in \mathbb{R}^F$ into a new vector of node features $\mathbf{h}'_v \in \mathbb{R}^{F'}$, where $F$ and $F'$ are the number of input and output features in each node, respectively.
We start by computing the attention coefficients in the neighborhood of node $v$ as follows,
\begin{equation}
  \alpha_{vu}^k = \underset{u\in \mathcal{N}_v\cup v}{\operatorname{softmax}}
    \left ( \operatorname{leakyrelu}({\mathbf{a}_k}^\intercal[\mathbf{W}_k \mathbf{h}_v + \mathbf{U}_k \mathbf{p}_v \| \mathbf{W}_k \mathbf{h}_u + \mathbf{U}_k \mathbf{p}_u])\right)
\end{equation}
where $\mathbf{W}_k$, $\mathbf{U}_k$ and $\mathbf{a}_k$ are the weights in the $k$-th attention head; $\mathbf{p}_v$ is the positional embedding for node $v$; $\|$ denotes the concatenation; and \[\underset{u\in \mathcal{U}}{\operatorname{softmax}}(e_{vu})= \frac{\exp(e_{vu})}{\sum_{u'\in \mathcal{U}}\exp(e_{vu'})}\]

The attention coefficients, computed based on the input node features and structural information, are expected to exploit the structural and semantic information within the neighborhood. We subsequently update the features of node $v$ by linear transforming the features of the nodes in the neighborhood with the obtained attention coefficients,
\begin{equation}
\begin{aligned}
     \mathbf{h}'_v  &= \left\{ 
     \begin{array}{cc}
      \|_{k=1}^K \sigma(\sum_{u\in \mathcal{N}_v \cup v} \alpha_{vu}^k \cdot \mathbf{W}_k {\mathbf{h}}_u)   &, \text{ if at the hidden layers}  , \\ 
   \sigma\left(\frac{1}{K}\sum_{k=1}^K\sum_{u\in \mathcal{N}_v \cup v} \alpha_{vu}^k  \cdot \mathbf{W}_k {\mathbf{h}}_u\right)   &,  \text{if at the output layer}.
     \end{array} 
     \right.
     \end{aligned}
\end{equation}
where $\sigma$ denotes nonlinear activation function. Consequently, the features extracted by each of the positional embedding enhanced attentional layers should be able to explore the structural and semantic information in each neighborhood.

Finally, the enhanced GATs in GAT-POS is constructed by stacking multiple such graph attentional layers.
In the first layer, $\mathbf{h}_v$ is set as $\mathbf{x}_v$ and we denote the output of the final layer as the predicted node label $\hat{\mathbf{y}}_v$.

\subsection{Positional Embedding Model}
We propose to learn the positional embeddings via an embedding model with an unsupervised task. In order to guide the positional embedding model to learn positional and structural information of the graph, we employ the unsupervised objective function utilized in many graph embedding models~\cite{tang2015line,hamilton2017inductive,yang2016revisiting} based on the skip-gram model~\cite{mikolov2013distributed}.
This objective function is a computationally efficient approximation to the cross-entropy error of predicting first-order and second-order proximities among nodes, via a negative sampling scheme.
More precisely, we define
\begin{equation}
    \mathcal{L}_{\mathcal{U}}(\{\mathbf{p}_v\}_{v\in \mathcal{V}},G)) = \sum_{v \in \mathcal{V}} \sum_{u \in \mathcal{N}(v)} \left( - \log \sigma({\mathbf{p}_v}^\intercal \mathbf{p}_u) - Q \cdot \mathbb{E}_{u'\sim P_n(v)}\log (\sigma(-{\mathbf{p}_v}^\intercal \mathbf{p}_{u'}))
    \right),
\end{equation}
where $P_n(v)$ denotes the distribution negative sampling for node $v$; and $Q$ is the number of negative samples per edge.

With the unsupervised objective, we utilize a positional embedding model with simple architecture, which is constructed by stacking multiple fully-connected layers.
The $t$-th layer of the positional embedding model is computed as follows,
\begin{equation}
    \mathbf{p}^{t}_v = \sigma(\mathbf{W}_{emb}^t \mathbf{p}^{t-1}_v)
\end{equation}
where $\mathbf{p}^t_v$ and $\mathbf{W}_{emb}^t$ are the learned positional representation for node $v$ and the weight matrix at layer $t$, respectively; and $\sigma$ denotes an arbitrary nonlinear activation.
Even though more complicated embedding models may be used, embedding models with such simple architectures can still capture meaningful information of the graph structure according to previous works~\cite{mikolov2013distributed,yang2016revisiting,tang2015line}.
We only consider the transductive positional embedding model given the nature of the datasets used in our experiments.
Thus, $\mathbf{p}^0_v$ is the learned initial positional embedding for node $v$ and $\mathbf{p}_v$ is the corresponding output positional embedding from the final layer.

\section{Related Works}
\label{sec:related_work}
\subsubsection{Attention-based Models}
Attention mechanisms have been widely used in many sequence-based and vision-based tasks.

Attention mechanisms were firstly proposed in the machine translation literature to overcome the information bottleneck problem of encoder-decoder RNN-based architectures \cite{bahdanau2014neural,luong2015effective}. Similar models were designed to enhance image captioning architectures \cite{xu2015show}.
Since then, self-attention and attention-based models have been extensively utilized as feature extractors, which allow for variable-sized inputs~\cite{vaswani2017attention,shaw2018self,dai2019transformerxl,wang2019structuralpos,cordonnier2019relationship,zhao2020exploring,velivckovic2017graph}. Among those, Transformers~\cite{vaswani2017attention} and GATs~\cite{velivckovic2017graph} are closely related to our proposed framework.
The former is a well-known language model based on attention mechanisms exclusively.
Particularly, self-attention is utilized to learn word representations in a sentence, capturing their syntactic and semantic information.
Note that self-attention assumes the words under the set structure, regardless their associated structural dependencies.
Therefore, to explore the positional information of words in a sentence, a positional encoding was introduced in the Transformer and its variants. Following the Transformer, GATs learn node reprensentations in a graph with self-attention.
More specifically, GATs mask out the interactions between unconnected nodes to somehow make use of the graph structure. This simple masking technique allows GATs to capture the co-occurrence of nodes in each neighborhood but, unfortunately, does not enable the model to fully explore the structure of the graph.
Thus, we propose to learn positional embeddings to enhance the ability of GATs to fully explore the structural information of a graph.
Consequently, our model, GAT-POS, can learn node representations which better capture the syntactic and semantic information in the graph.

\subsubsection{Other Structural Embedding Enhanced GNNs}
As pointed by \cite{pei2020geom}, GNNs struggle to fully explore the structural information of a graph, and this shortcoming limits their effective application to non-homophilic graphs, which require increased understanding of higher-level structural dependencies. 
To remedy this, position-aware Graph Neural Networks (PGNNs)~\cite{you2019position} introduced the concept of position-aware node embeddings for pairwise node classification and link prediction.
In order to capture the position or location information of a given target node, in addition to the architecture of vanilla GNNs, PGNN samples sets of anchor nodes, computes the distances of the target node to each anchor-set, and learns a non-linear aggregation scheme for distances over the anchor-sets.
It is worth mentioning that the learned position-aware embeddings are permutation sensitive to the order of nodes in the anchor-sets. Moreover, to learn on non-homophilic graphs, \cite{pei2020geom} proposes a framework, termed Geom-GCN, in which GNNs are enhanced with latent space embeddings, capturing structural information of the graph.
As opposed to GAT-POS,
Geom-GCN does not support end-to-end training, and therefore their embeddings cannot be adaptively adjusted for different supervised tasks.
Similar to GAT-POS, Geom-GCN also employs adaptive filters to extract features from each locality.
However, the adaptive filters in Geom-GCN are merely conditioned on the latent space embeddings, which do not leverage node content information, and might be less effective than our GAT-POS' attention mechanisms based on both node content and positional information.
Moreover, Geom-GCN also introduces an extended neighborhood that goes beyond the spatial neighborhood of each node, and which corresponds to the set of nodes close to the center node in the latent space.
However, according to the experimental results of our paper, the extended neighborhood of the Geom-GCN is not beneficial when dealing with undirected graphs. 
Finally, a very recent concurrent work~\cite{li2020distance} enhances GNNs with distance encodings for GNNs. 
As opposed to Geom-GCN and GAT-POS, the distance encoding is built without any training process on the graph, capturing graph properties such as shortest-path-distance and generalized PageRank scores based on an encoding process.

\vspace{0.5cm}

\section{Experiments and Results}

We have performed comparative evaluation experiments of GAT-POS on non-homophilic graph-structured datasets against standard GNNs (i.e., GCNs and GATs) and related embedding enhanced GNNs (i.e., Geom-GCN).
Our proposed framework reaches remarkably improved performance compared to standard GNNs and outperforms Geom-GCNs.
Besides, we also performed a sanity check test on homophilic datasets to validate that GAT-POS can reach comparable performance to standard GNNs.

\subsection{Datasets}

We utilize two Wikipedia page-page networks (i.e., Chameleon and Squirrel)~\cite{rozemberczki2021multi} and the Actor co-occurrence network (shortened as Actor) introduced in~\cite{pei2020geom} as our non-homophilic datasets.
We also include three widely used citation networks (i.e., Cora, Citeseer and Pubmed)~\cite{yang2016revisiting,kipf2016semi,velivckovic2017graph} as representatives of homophilic datasets for our sanity check. Note that all graphs are converted to undirected graphs, following the widely utilized setup in other works~\cite{kipf2016semi,velivckovic2017graph}.

Table~\ref{tab:data_stat} summarizes basic statistics of each dataset, together with their homophily levels. The homophily level of a graph dataset can be demonstrated by a measure $\beta$ introduced in ~\cite{pei2020geom}, which computes the ratio of the nodes in each neighborhood sharing the same labels as the center node.

\begin{table}[t!]
\vspace{-0.5cm}
    \centering
    \begin{tabular}{l 
    >{\centering\arraybackslash}p{1.4cm} >{\centering\arraybackslash}p{1.4cm} >{\centering\arraybackslash}p{1.4cm} 
    >{\centering\arraybackslash}p{1.4cm} >{\centering\arraybackslash}p{1.4cm} >{\centering\arraybackslash}p{1.4cm}}
    \toprule
\multirow{2}{*}{\textbf{Dataset}} & \multicolumn{3}{c}{\textbf{Homophilic Datasets} } & \multicolumn{3}{c}{\textbf{Non-homophilic Datasets}}   \\ \cmidrule{2-4}\cmidrule(lr){5-7} 
 &  \textbf{Cora} & \textbf{Citeseer} & \textbf{Pubmed} & \textbf{Chameleon} & \textbf{Squirrel} & \textbf{Actor}\\ \midrule
       \textbf{Homophily}($\beta$)  &\cellcolor{green!20}  0.83 &\cellcolor{green!20} 0.71 &\cellcolor{green!20} 0.79 &\cellcolor{blue!20} 0.25 & \cellcolor{blue!20}0.22 & \cellcolor{blue!20}0.24 \\ \midrule
    \textbf{\#  Nodes} & 2708 & 3327 & 19717 & 2277 & 5201 & 7600  \\
    \textbf{\# Edges\textsuperscript{*}} & 5429 & 4732 & 44338 & 36101 & 217073 & 33544 \\ 
    \textbf{\# Features} & 1433 & 3703 & 500 & 2325 & 2089 & 931 \\
    \textbf{\# Classes} & 7 & 6 & 3 & 5 & 5 & 5 \\ 
       \bottomrule
    \end{tabular}\\

    \caption{Summary of the datasets used in our experiments.} 
    \label{tab:data_stat}
    \vspace{-0.8cm}
\end{table}

Following the evaluation setup in ~\cite{pei2020geom}, for all graph datasets, we randomly split nodes of each class into $60\%$, $20\%$, and $20\%$ for training, validation and testing, respectively.
The final experimental results reported are the average classification accuracies of all models on the test sets over $10$ random splits for all graph datasets.
For each split, 10 runs of experiments are performed with different random initializations.
We utilize the splits provided by~\cite{pei2020geom}.
Note that all datasets in our evaluation are under the transductive setting.

\subsection{Methods in Comparison}
In order to demonstrate that our framework can effectively enhance the performance of GATs on non-homophilic datasets, the standard GAT~\cite{velivckovic2017graph} is considered in the comparison.
Furthermore, another commmonly used GNN, the GCN~\cite{kipf2016semi} is also included in our evaluation as a representative of spectral GNNs. 
We also include Geom-GCN~\cite{pei2020geom} into our evaluation as a baseline of structural embedding enhanced GNNs.

Specifically, three variants of Geom-GCNs coupled with different embedding methods are considered, namely Geom-GCN with Isomap (Geom-GCN-I), Geom-GCN with Poincare embedding (Geom-GCN-P), and Geom-GCN with struc2vec embedding (Geom-GCN-S).
Besides, we also include the variants of Geom-GCNs without their extended neighborhood scheme, considering only spatial neighborhoods in the graph, termed Geom-GCN-I-g, Geom-GCN-P-g and Geom-GCN-S-g respectively.

\subsection{Experimental Setup}
We utilize the hyperparameter settings of baseline models from~\cite{pei2020geom} in our experiments since they have already undergone extensive hyperparameter search on each dataset's validation set. 
For our proposed GAT-POS, we perform a hyperparameter search on the number of hidden units, the initial learning rate, the weight decay, and the probability of dropout.
For fairness, the hyperparameter search is performed on the same searching space as the models tuned in ~\cite{pei2020geom}.
Specifically, the number of layers of GNN architectures is fixed to $2$ and the Adam optimizer \cite{kingma2014adam} is used to train all models.

For GAT-POS, according to the result of the hyperparameter search, 
we set the initial learning rate to $5\text{e-}3$, weight decay to $5\text{e-}4$, a dropout of $p=0.5$ and the number of hidden units for the positional embedding model to $64$.
Besides, the activation functions of the positional embedding model and the enhanced GAT architecture are set to ReLU and ELU, respectively.
The number of hidden units per attention head in the main architecture is $8$ (Cora, Citeseer, Pubmed and Squirrel) and $32$ (Chameleon and Actor).
The number of attention heads for the hidden layer is $8$ (Cora, Citeseer and Pubmed) and $16$ (Chameleon, Squirrel and Actor).
For all datasets, the number of attention heads in the output layer is $1$ and the residual connections are employed.

\begin{table}[b!]
\vspace{-0.5cm}
\centering
\begin{tabular}{>{\arraybackslash}p{2.5cm}
>{\centering\arraybackslash}p{3cm}>{\centering\arraybackslash}p{3cm}>{\centering\arraybackslash}p{3cm}}
\toprule
{} & \multicolumn{3}{c}{\cellcolor{green!20}\textbf{Non-Homophilic Datasets}} \\
\textbf{Method} &       \textbf{Chameleon} &         \textbf{Squirrel} &            \textbf{Actor} \\
\toprule
GCN            & $  65.22 \pm 2.22\%$ & $  45.44 \pm 1.27\%$ & $  28.30 \pm 0.73\%$ \\
\cellcolor{gray!20} GAT            & \cellcolor{gray!20}$  63.88 \pm 2.42\%$ &\cellcolor{gray!20} $  41.19 \pm 3.38\%$ & \cellcolor{gray!20}$  28.49 \pm 1.06\%$ \\ \midrule
Geom-GCN-I      & $  57.35 \pm 1.85\%$ & $  31.92 \pm 1.04\%$ & $  29.14 \pm 1.15\%$ \\
Geom-GCN-P      & $  60.68 \pm 1.97\%$ & $  35.39 \pm 1.21\%$ & $  31.92 \pm 0.95\%$ \\
Geom-GCN-S      & $  57.89 \pm 1.65\%$ & $  35.74 \pm 1.47\%$ & $  30.12 \pm 0.92\%$ \\
Geom-GCN-I-g    &  $61.97 \pm 2.01\% $ &  $39.91 \pm 1.77\% $ &  $32.98 \pm 0.78\% $ \\
Geom-GCN-P-g    &  $60.31 \pm 2.07\% $ &  $37.14 \pm 1.19\% $ &  $31.60 \pm 1.06\% $ \\
Geom-GCN-S-g    &  $63.86 \pm 1.78\% $ &  $42.73 \pm 2.16\% $ &  $31.96 \pm 1.97\% $ \\
\midrule
\cellcolor{gray!20} GAT-POS (ours)   & \cellcolor{gray!20}$  \mathbf{67.76 \pm 2.54\%}$ &\cellcolor{gray!20} $  \mathbf{52.90 \pm 1.55\%}$ &\cellcolor{gray!20} $  \mathbf{34.89 \pm 1.38\%}$ \\
\bottomrule
\end{tabular} \\
 \caption{The proposed positional embedding aware models outperforms their corresponding baselines in terms of accuracy on non-homophilic datasets and GAT-POS shows significantly better performance compared to Geom-GCN variations. 
}
\label{tab:result_nonhomo}
\end{table}

\subsection{Results}
The results of our comparative evaluation experiments on non-homophilic and homophilic datasets are summarized in Tables~\ref{tab:result_nonhomo} and \ref{tab:result_homo}, respectively.

\subsubsection{Non-homophilic Graph Datasets}
The experimental results demonstrate that our framework reaches better performance than baselines on all non-homophilic graphs.
Compared to GATs, GAT-POS achieves a remarkable improvement on the three non-homophilic datasets.
Our proposed framework also reaches a better performance compared to GCNs and Geom-GCNs.
Unexpectedly, Geom-GCNs does not show significant improvement compared to standard GNNs on undirected graphs.
This is against the previous experimental results on directed graphs.
One possible reason is that the Geom-GCNs' extended neighborhood scheme cannot include extra useful neighbors in the aggregation but rather introduces redundant operations and parameters to train on undirected graphs.
This is also demonstrated by the better results of the instantiations of Geom-GCNs with only spatial neighborhoods.

\subsubsection{Homophilic Graph Datasets}

Recall that the evaluation of homophilic datasets is only performed as a sanity check, since GAT-POS has been tailored to exploit structural and positional information essential to non-homophilic tasks. As shown in the table, our proposed method reaches comparable results to standard GNNs.
 
These observations are in line with our expectations on homophilic datasets since merely learning underlying group-invariances can already lead to good performance on graph datasets with high homophily.

\begin{table}[b!]
\vspace{-0.5cm}
\centering
\begin{tabular}{>{\arraybackslash}p{2.5cm}
>{\centering\arraybackslash}p{3cm}>{\centering\arraybackslash}p{3cm}>{\centering\arraybackslash}p{3cm}}
\toprule
{} & \multicolumn{3}{c}{\cellcolor{blue!20}\textbf{Homophilic Datasets}} \\
\textbf{Method} &             \textbf{Cora} &         \textbf{Citeseer} &           \textbf{Pubmed} \\
\toprule
GCN            & $  85.67 \pm 0.94\%$ & $  73.28 \pm 1.37\%$ & $  88.14 \pm 0.32\%$ \\
\cellcolor{gray!20}GAT            &\cellcolor{gray!20} $ \mathbf{87.06 \pm 0.98\%}$ &\cellcolor{gray!20} $  74.79 \pm 1.89\%$ &\cellcolor{gray!20} $  87.51 \pm 0.43\%$ \\ \midrule
Geom-GCN-I      & $  84.79 \pm 2.04\%$ & $ { 78.84 \pm 1.51\%}$ & $  {89.73 \pm 0.54\%}$ \\
Geom-GCN-P      & $  84.68 \pm 1.59\%$ & $  73.77 \pm 1.59\%$ & $  88.15 \pm 0.57\%$ \\
Geom-GCN-S      & $  85.25 \pm 1.46\%$ & $  74.42 \pm 2.52\%$ & $  84.80 \pm 0.62\%$ \\
Geom-GCN-I-g    &  $85.81 \pm 1.50\% $ &  $\mathbf{80.05 \pm 1.59\%} $ &  $\mathbf{92.56 \pm 0.33\% }$ \\
Geom-GCN-P-g    &  $86.48 \pm 1.49\% $ &  $75.74 \pm 1.58\% $ &  $88.49 \pm 0.56\% $ \\
Geom-GCN-S-g    &  $86.50 \pm 1.43\% $ &  $75.91 \pm 2.26\% $ &  $88.55 \pm 0.53\% $ \\
\midrule
\cellcolor{gray!20}GAT-POS (ours)   &\cellcolor{gray!20} $  86.61 \pm 1.13\%$ &\cellcolor{gray!20} $  73.81 \pm 1.27\%$ &\cellcolor{gray!20} $ 87.56 \pm 0.48\% $ \\
\bottomrule
\end{tabular}\\
\caption{The proposed positional embedding aware models are on par with their corresponding baselines in terms of accuracy on homophilic datasets; in particular, GAT-POS's results are comparable to GATs.} 
\label{tab:result_homo}
\vspace{-0.8cm}
\end{table}

\subsection{Ablation Study}

In this section, we present an ablation study to understand the contributions of the join-training scheme. To that end, we include a variant of GAT-POS without joint-training. This variant pretrains the positional embedding model with the same unsupervised task and then freezes the learnt embeddings while training the enhanced GATs with the supervised objective.

Our proposed implementation of the enhanced GAT is inspired the work applying attention mechanism on images~\cite{zhao2020exploring}.
Thus, we also introduce a variant following the architecture of Transformer from natural language processing domain, termed GAT-POS-Transformer.
The difference between both architectures lies in that GAT-POS only considers the positional embeddings in the computation of attention coefficients but not in the neighborhood aggregation, while
GAT-POS-Transformer directly injects the positional embeddings into the node features before feeding them to the attention module.
Note that, GAT-POS-Transformer utilizes the architecture of the standard GATs since the input features have been enhanced by the positional embeddings prior to the attention computation. The results of the comparison are summarized in Table~\ref{tab:ablation}.

\begin{table}[b!]
\vspace{-0.5cm}
\centering
\begin{tabular}{>{\arraybackslash}p{2.5cm} c c c c }
\toprule
 
{} &   \multicolumn{2}{c}{ GAT-POS}  & \multicolumn{2}{c}{GAT-POS-Transformer}  \\ 
\cmidrule{2-3}\cmidrule{4-5} 
Joint-Training & \ding{51} & \ding{55} & \ding{51} & \ding{55}  \\ \midrule 
Chameleon &     $  \mathbf{67.76 \pm 2.54\%}$ & $ 65.75 \pm 1.81\% $ &   $ \mathbf{ 65.55 \pm 2.38\%}$ & $ 65.42 \pm 2.13\% $ \\
Squirrel & $  \mathbf{52.90 \pm 1.55\%}$ & $ 50.63 \pm 1.29\% $ &  $  \mathbf{51.62 \pm 1.84\%}$ & $ 50.79 \pm 1.35\% $ \\
Actor &  $  34.89 \pm 1.38\%$ &  $ \mathbf{34.95 \pm 0.95\%} $ &  $ \mathbf{ 34.97 \pm 1.27\%}$ &  $ 34.66 \pm 1.17\% $ \\
\bottomrule
\end{tabular}
\begin{adjustwidth}{1cm}{1cm}
\end{adjustwidth}
\caption{The Ablation Study of GAT-POS}
\label{tab:ablation}
\vspace{-0.8cm}
\end{table}

On Chameleon and Squirrel, both instantiations of GAT-POS with joint-training reach better results than without joint-training on average.
However, it is worth noting that the joint-training also leads to relatively larger standard deviations.
On Actor datasets, the difference of performances between the instantiations with and without joint-training is not notable.
This might be because, on Actor dataset, the positional embedding is probably less affected by the supervised signals.
Overall, enabling joint training of the main model and the positional embedding model is beneficial. Moreover, it is worth noting that the variants of GAT-POS without joint-training still outperform Geom-GCNs by a notable margin, which also highlights the advantage of having an attention mechanism incorporate semantic and structural information, compared to the weighting functions in Geom-GCNs, which only considers the structural information.

In the comparison between GAT-POS and GAT-POS-Transformer, GAT-POS reaches slightly better performance than GAT-POS-Transformer on average, but the difference is not significant.
A deeper research on the architecture design of the enhanced GATs will be left for the future work.

\section{Conclusion}

We have presented a framework to enhance the GAT models with a positional embedding to explicitly leverage both the semantic information of each node and the structural information contained in the graph. 
In particular, we extended the standard GAT formulation by adding a positional embedding model, predictive of the graph context, and connecting it to the enhanced GAT model through the proposed attentional layer. 
Although we focused on extending the original GAT formulation, our proposed framework is compatible with most current graph deep learning based models, which leverage graph attentional layers. 
Moreover, it is worth mentioning that this framework is agnostic to the choice of positional embedding models as well as the particular setup of the graph attentional layers.

Experiments on a number of graph-structured datasets, especially those with more complicated structures and edges joining nodes beyond simple feature similarity, suggest that this framework can effectively enhance GATs by leveraging both graph structure and node attributes for node classification tasks, resulting in increased performances when compared to baselines as well as recently introduced methods. Finally, through an ablation study, we have further emphasized the benefits of our proposed framework by showing that the performance improvements do not only come from the joint training of both parts of the model but rather from endowing GATs with node positional information.

There are several potential improvements to our proposed framework that could be addressed as future work.
One is about improving the generalization ability of GAT-POS in the inductive setting.
Even though GAT-POS can support the inductive setting by utilizing inductive embedding methods such as GraphSAGE, the bottleneck is still on the embedding methods, especially when the input node features are homogeneous.
Another potential research direction is asynchronous training on supervised and unsupervised tasks with the overall architecture, which might be essential when learning on super large-scale graph.


\bibliographystyle{splncs04}
\bibliography{ref}
\end{document}